\PassOptionsToPackage{unicode}{hyperref}
\PassOptionsToPackage{hyphens}{url}
\PassOptionsToPackage{dvipsnames,svgnames,x11names}{xcolor}
\documentclass[
]{article}
\usepackage{amsmath,amssymb}
\usepackage{lmodern}
\usepackage{iftex}
\usepackage[T1]{fontenc}
\usepackage[utf8]{inputenc}
\usepackage{textcomp} 
\IfFileExists{upquote.sty}{\usepackage{upquote}}{}
\IfFileExists{microtype.sty}{
    \usepackage[]{microtype}
    \UseMicrotypeSet[protrusion]{basicmath} 
}{}
\makeatletter
\@ifundefined{KOMAClassName}{
    \IfFileExists{parskip.sty}{%
        \usepackage{parskip}
    }{
        \setlength{\parindent}{0pt}
        \setlength{\parskip}{6pt plus 2pt minus 1pt}}
}{
    \KOMAoptions{parskip=half}}
\makeatother
\usepackage{xcolor}
\usepackage{graphicx}
\makeatletter
\def\maxwidth{\ifdim
                  \Gin@nat@width>\linewidth\linewidth\else
                  \Gin@nat@width
\fi}
\def\maxheight{\ifdim
                   \Gin@nat@height>\textheight\textheight\else
                   \Gin@nat@height
\fi}
\makeatother
\setkeys{Gin}{width=\maxwidth,height=\maxheight,keepaspectratio}
\makeatletter
\def\fps@figure{htbp}
\makeatother
\providecommand{\tightlist}{%
    \setlength{\itemsep}{0pt}\setlength{\parskip}{0pt}}
\NewDocumentCommand\citeproctext{}{}
\NewDocumentCommand\citeproc{mm}{%
    \begingroup\def\citeproctext{#2}\cite{#1}\endgroup}
\makeatletter
\let\@cite@ofmt\@firstofone
\def\@biblabel#1{}
\def\@cite#1#2{{#1\if@tempswa , #2\fi}}
\makeatother
\newlength{\cslhangindent}
\newlength{\csllabelwidth}
\newenvironment{CSLReferences}[2] 
{\begin{list}{}{%
    \setlength{\itemindent}{0pt}
    \setlength{\leftmargin}{0pt}
    \setlength{\parsep}{0pt}
    \ifodd
        #1
        \setlength{\leftmargin}{\cslhangindent}
        \setlength{\itemindent}{-1\cslhangindent}
    \fi
    \setlength{\itemsep}{#2\baselineskip}}}
{\end{list}}
\usepackage{calc}

\ifLuaTeX
\usepackage[bidi=basic]{babel}
\else
\usepackage[bidi=default]{babel}
\fi
\babelprovide[main,import]{american}

\def\languageshorthands#1{}
\ifLuaTeX
\usepackage{selnolig}  
\fi
\IfFileExists{bookmark.sty}{\usepackage{bookmark}}{\usepackage{hyperref}}
\IfFileExists{xurl.sty}{\usepackage{xurl}}{} 
\hypersetup{
    pdftitle={Causal identification with Y\_0},
    pdfauthor={Charles Tapley Hoyt, Craig Bakker, Richard J. Callahan,
    Joseph Cottam, August George, Benjamin M. Gyori, Haley M. Hummel,
    Nathaniel Merrill, Sara Mohammad Taheri, Pruthvi Prakash Navada,
    Marc-Antoine Parent, Adam Rupe, Olga Vitek, Jeremy Zucker},
    pdflang={en-US},
    colorlinks=true,
    linkcolor={Maroon},
    filecolor={Maroon},
    citecolor={Blue},
    urlcolor={Blue},
    pdfcreator={LaTeX via pandoc}}

\title{Causal identification with \(Y_0\)}

\definecolor{c53baa1}{RGB}{83,186,161}
\definecolor{c202826}{RGB}{32,40,38}
\def \rorglobalscale {0.1}
\newcommand{\rorlogo}{%
    \begin{tikzpicture}[y=1cm, x=1cm, yscale=\rorglobalscale,xscale=\rorglobalscale, every node/.append style={scale=\rorglobalscale}, inner sep=0pt, outer sep=0pt]
    \begin{scope}[even odd rule,line join=round,miter limit=2.0,shift={(-0.025, 0.0216)}]
    \path[fill=c53baa1,nonzero rule,line join=round,miter limit=2.0] (1.8164, 3.012) -- (1.4954, 2.5204) -- (1.1742, 3.012) -- (1.8164, 3.012) -- cycle;
    \path[fill=c53baa1,nonzero rule,line join=round,miter limit=2.0] (3.1594, 3.012) -- (2.8385, 2.5204) -- (2.5172, 3.012) -- (3.1594, 3.012) -- cycle;
    \path[fill=c53baa1,nonzero rule,line join=round,miter limit=2.0] (1.1742, 0.0669) -- (1.4954, 0.5588) -- (1.8164, 0.0669) -- (1.1742, 0.0669) -- cycle;
    \path[fill=c53baa1,nonzero rule,line join=round,miter limit=2.0] (2.5172, 0.0669) -- (2.8385, 0.5588) -- (3.1594, 0.0669) -- (2.5172, 0.0669) -- cycle;
    \path[fill=c202826,nonzero rule,line join=round,miter limit=2.0] (3.8505, 1.4364).. controls (3.9643, 1.4576) and (4.0508, 1.5081) .. (4.1098, 1.5878).. controls (4.169, 1.6674) and (4.1984, 1.7642) .. (4.1984, 1.8777).. controls (4.1984, 1.9719) and (4.182, 2.0503) .. (4.1495, 2.1132).. controls (4.1169, 2.1762) and (4.0727, 2.2262) .. (4.0174, 2.2635).. controls (3.9621, 2.3006) and (3.8976, 2.3273) .. (3.824, 2.3432).. controls (3.7505, 2.359) and (3.6727, 2.367) .. (3.5909, 2.367) -- (2.9676, 2.367) -- (2.9676, 1.8688).. controls (2.9625, 1.8833) and (2.9572, 1.8976) .. (2.9514, 1.9119).. controls (2.9083, 2.0164) and (2.848, 2.1056) .. (2.7705, 2.1791).. controls (2.6929, 2.2527) and (2.6014, 2.3093) .. (2.495, 2.3487).. controls (2.3889, 2.3881) and (2.2728, 2.408) .. (2.1468, 2.408).. controls (2.0209, 2.408) and (1.905, 2.3881) .. (1.7986, 2.3487).. controls (1.6925, 2.3093) and (1.6007, 2.2527) .. (1.5232, 2.1791).. controls (1.4539, 2.1132) and (1.3983, 2.0346) .. (1.3565, 1.9436).. controls (1.3504, 2.009) and (1.3351, 2.0656) .. (1.3105, 2.1132).. controls (1.2779, 2.1762) and (1.2338, 2.2262) .. (1.1785, 2.2635).. controls (1.1232, 2.3006) and (1.0586, 2.3273) .. (0.985, 2.3432).. controls (0.9115, 2.359) and (0.8337, 2.367) .. (0.7519, 2.367) -- (0.1289, 2.367) -- (0.1289, 0.7562) -- (0.4837, 0.7562) -- (0.4837, 1.4002) -- (0.6588, 1.4002) -- (0.9956, 0.7562) -- (1.4211, 0.7562) -- (1.0118, 1.4364).. controls (1.1255, 1.4576) and (1.2121, 1.5081) .. (1.2711, 1.5878).. controls (1.2737, 1.5915) and (1.2761, 1.5954) .. (1.2787, 1.5991).. controls (1.2782, 1.5867) and (1.2779, 1.5743) .. (1.2779, 1.5616).. controls (1.2779, 1.4327) and (1.2996, 1.3158) .. (1.3428, 1.2113).. controls (1.3859, 1.1068) and (1.4462, 1.0176) .. (1.5237, 0.944).. controls (1.601, 0.8705) and (1.6928, 0.8139) .. (1.7992, 0.7744).. controls (1.9053, 0.735) and (2.0214, 0.7152) .. (2.1474, 0.7152).. controls (2.2733, 0.7152) and (2.3892, 0.735) .. (2.4956, 0.7744).. controls (2.6016, 0.8139) and (2.6935, 0.8705) .. (2.771, 0.944).. controls (2.8482, 1.0176) and (2.9086, 1.1068) .. (2.952, 1.2113).. controls (2.9578, 1.2253) and (2.9631, 1.2398) .. (2.9681, 1.2544) -- (2.9681, 0.7562) -- (3.3229, 0.7562) -- (3.3229, 1.4002) -- (3.4981, 1.4002) -- (3.8349, 0.7562) -- (4.2603, 0.7562) -- (3.8505, 1.4364) -- cycle(0.9628, 1.7777).. controls (0.9438, 1.7534) and (0.92, 1.7357) .. (0.8911, 1.7243).. controls (0.8623, 1.7129) and (0.83, 1.706) .. (0.7945, 1.7039).. controls (0.7588, 1.7015) and (0.7252, 1.7005) .. (0.6932, 1.7005) -- (0.4839, 1.7005) -- (0.4839, 2.0667) -- (0.716, 2.0667).. controls (0.7477, 2.0667) and (0.7805, 2.0643) .. (0.8139, 2.0598).. controls (0.8472, 2.0553) and (0.8768, 2.0466) .. (0.9025, 2.0336).. controls (0.9282, 2.0206) and (0.9496, 2.0021) .. (0.9663, 1.9778).. controls (0.9829, 1.9534) and (0.9914, 1.9209) .. (0.9914, 1.8799).. controls (0.9914, 1.8362) and (0.9819, 1.8021) .. (0.9628, 1.7777) -- cycle(2.6125, 1.3533).. controls (2.5889, 1.2904) and (2.5553, 1.2359) .. (2.5112, 1.1896).. controls (2.4672, 1.1433) and (2.4146, 1.1073) .. (2.3529, 1.0814).. controls (2.2916, 1.0554) and (2.2228, 1.0427) .. (2.1471, 1.0427).. controls (2.0712, 1.0427) and (2.0026, 1.0557) .. (1.9412, 1.0814).. controls (1.8799, 1.107) and (1.8272, 1.1433) .. (1.783, 1.1896).. controls (1.7391, 1.2359) and (1.7052, 1.2904) .. (1.6817, 1.3533).. controls (1.6581, 1.4163) and (1.6465, 1.4856) .. (1.6465, 1.5616).. controls (1.6465, 1.6359) and (1.6581, 1.705) .. (1.6817, 1.7687).. controls (1.7052, 1.8325) and (1.7388, 1.8873) .. (1.783, 1.9336).. controls (1.8269, 1.9799) and (1.8796, 2.0159) .. (1.9412, 2.0418).. controls (2.0026, 2.0675) and (2.0712, 2.0804) .. (2.1471, 2.0804).. controls (2.223, 2.0804) and (2.2916, 2.0675) .. (2.3529, 2.0418).. controls (2.4143, 2.0161) and (2.467, 1.9799) .. (2.5112, 1.9336).. controls (2.5551, 1.8873) and (2.5889, 1.8322) .. (2.6125, 1.7687).. controls (2.636, 1.705) and (2.6477, 1.6359) .. (2.6477, 1.5616).. controls (2.6477, 1.4856) and (2.636, 1.4163) .. (2.6125, 1.3533) -- cycle(3.8015, 1.7777).. controls (3.7825, 1.7534) and (3.7587, 1.7357) .. (3.7298, 1.7243).. controls (3.701, 1.7129) and (3.6687, 1.706) .. (3.6333, 1.7039).. controls (3.5975, 1.7015) and (3.5639, 1.7005) .. (3.5319, 1.7005) -- (3.3226, 1.7005) -- (3.3226, 2.0667) -- (3.5547, 2.0667).. controls (3.5864, 2.0667) and (3.6192, 2.0643) .. (3.6526, 2.0598).. controls (3.6859, 2.0553) and (3.7155, 2.0466) .. (3.7412, 2.0336).. controls (3.7669, 2.0206) and (3.7883, 2.0021) .. (3.805, 1.9778).. controls (3.8216, 1.9534) and (3.8301, 1.9209) .. (3.8301, 1.8799).. controls (3.8301, 1.8362) and (3.8206, 1.8021) .. (3.8015, 1.7777) -- cycle;
    \end{scope}
    \end{tikzpicture}
}

\usepackage[affil-it]{authblk}
\usepackage{orcidlink}
\author[1%
*%
\ensuremath\mathparagraph]{Charles Tapley Hoyt%
\,\orcidlink{0000-0003-4423-4370}\,%
}
\author[2%
]{Craig Bakker%
\,\orcidlink{0000-0002-0083-4000}\,%
}
\author[2%
]{Richard J. Callahan%
\,\orcidlink{0009-0006-6041-5517}\,%
}
\author[2%
]{Joseph Cottam%
\,\orcidlink{0000-0002-3097-5998}\,%
}
\author[2%
]{August George%
\,\orcidlink{0000-0001-7876-4359}\,%
}
\author[3%
]{Benjamin M. Gyori%
\,\orcidlink{0000-0001-9439-5346}\,%
}
\author[2,4%
]{Haley M. Hummel%
\,\orcidlink{0009-0004-5405-946X}\,%
}
\author[5%
]{Nathaniel Merrill%
\,\orcidlink{0000-0002-1998-0980}\,%
}
\author[3%
]{Sara Mohammad Taheri%
\,\orcidlink{0000-0002-6554-9083}\,%
}
\author[3%
]{Pruthvi Prakash Navada%
\,\orcidlink{0009-0008-8505-1670}\,%
}
\author[6%
]{Marc-Antoine Parent%
\,\orcidlink{0000-0003-4159-7678}\,%
}
\author[2%
]{Adam Rupe%
\,\orcidlink{0000-0003-0105-8987}\,%
}
\author[3%
]{Olga Vitek%
\,\orcidlink{0000-0003-1728-1104}\,%
}
\author[2%
*%
\ensuremath\mathparagraph]{Jeremy Zucker%
\,\orcidlink{0000-0002-7276-9009}\,%
}

\affil[1]{RWTH Aachen University%
\,\protect\href{https://ror.org/04xfq0f34}{\protect\rorlogo}\,%
}
\affil[2]{Pacific Northwest National Laboratory%
\,\protect\href{https://ror.org/05h992307}{\protect\rorlogo}\,%
}
\affil[3]{Northeastern University%
\,\protect\href{https://ror.org/04t5xt781}{\protect\rorlogo}\,%
}
\affil[4]{Oregon State University%
\,\protect\href{https://ror.org/00ysfqy60}{\protect\rorlogo}\,%
}
\affil[5]{Battelle Memorial Institute%
\,\protect\href{https://ror.org/01h5tnr73}{\protect\rorlogo}\,%
}
\affil[6]{Conversence%
}
\affil[$\mathparagraph$]{Corresponding author: %
cthoyt@gmail.com jeremy.zucker@pnnl.gov %
}
\affil[*]{These authors contributed equally.}
\date{21 June 2025}

\begin{document}
    \maketitle

    \section{Summary}\label{summary}

    Researchers are often interested in investigating whether one thing
    causes another, such as whether a medication effectively treats a
    disease or whether education improves income. Randomized controlled
    trials provide the most direct evidence for causal relationships, but
    they are often logistically impossible, unethical, or prohibitively
    expensive to conduct. Causal inference comprises statistical methods
    that provide indirect evidence for causal relationships based on
    whatever data is available, whether it comes from a (randomized)
    controlled trial, an observational study, or a combination of both.
    However, both the qualitative and quantitative investigation of
    causation remains challenging in the presence of (unknown) confounding
    variables---a converse to the old adage that correlation does not imply
    causation.

    A key step in causal inference is \textbf{causal identification} during
    which it is determined whether it is theoretically possible to estimate
    a causal effect from available data, given prior knowledge about
    relationships between variables and a causal query, such as:

    \begin{enumerate}
        \def\labelenumi{\arabic{enumi}.}
        \tightlist
        \item
        \textbf{Interventional Query}, which asks: \emph{what will happen if
        we intervene?} For example, what would be the average effect if
        everyone received treatment?
        \item
        \textbf{Counterfactual Query}, which asks: \emph{what would have
        happened had we done something different?} For example, would a given
        patient, who recovered after receiving treatment, have recovered
        anyway without treatment?.
        \item
        \textbf{Transportability Query}, which asks: \emph{can causal findings
        from one population be validly applied to another, and if so, how can
        evidence from multiple studies or populations be combined to draw
        conclusions about a target group of interest?}
    \end{enumerate}

    We present the \(Y_0\) Python package, which implements causal
    identification algorithms that apply interventional, counterfactual, and
    transportability queries to data from (randomized) controlled trials,
    observational studies, or mixtures thereof. \(Y_0\) focuses on the
    qualitative investigation of causation, helping researchers determine
    \emph{whether} a causal relationship can be estimated from available
    data before attempting to estimate \emph{how strong} that relationship
    is. Furthermore, \(Y_0\) provides guidance on how to transform the
    causal query into a symbolic estimand that can be non-parametrically
    estimated from the available data. \(Y_0\) provides a domain-specific
    language for representing causal queries and estimands as symbolic
    probabilistic expressions, tools for representing causal graphical
    models with unobserved confounders, such as acyclic directed mixed
    graphs (ADMGs), and implementations of numerous identification
    algorithms from the recent causal inference literature.

    \section{Statement of Need}\label{statement-of-need}

    Several open source Python packages have implemented the simplest
    identification algorithm (\texttt{ID}) from Shpitser \& Pearl
    (\citeproc{ref-shpitser2006id}{2006}) including
    \href{https://gitlab.com/causal/ananke}{Ananke}
    (\citeproc{ref-lee2023ananke}{J. J. R. Lee et al., 2023}),
    \href{https://github.com/pgmpy/pgmpy}{pgmpy}
    (\citeproc{ref-ankan2015pgmpy}{Ankan \& Panda, 2015}),
    \href{https://github.com/py-why/dowhy}{DoWhy}
    (\citeproc{ref-sharma2020dowhy}{Sharma \& Kiciman, 2020}), and
    \href{https://github.com/pedemonte96/causaleffect}{causaleffect-py}
    (\citeproc{ref-pedemonte2021causalefffectpy}{Pedemonte et al., 2021}).
    Further, Ananke and DoWhy implement algorithms that consume the estimand
    returned by \texttt{ID} and observational data in order to estimate the
    average causal effect of an intervention on the outcome. However, these
    methods are limited in their generalization when causal queries include
    multiple outcomes, conditionals, or interventions.

    In the R programming language, the
    \href{https://github.com/santikka/causaleffect}{causaleffect} package
    (\citeproc{ref-tikka2017causaleffectr}{Tikka \& Karvanen, 2017a})
    implements \texttt{ID}, \texttt{IDC}
    (\citeproc{ref-shpitser2007idc}{Shpitser \& Pearl, 2008}), surrogate
    outcomes (\texttt{TRSO}) (\citeproc{ref-tikka2019trso}{Tikka \&
    Karvanen, 2019}), and transport (\citeproc{ref-correa2020transport}{S.
    Lee et al., 2020}). The \href{https://github.com/santikka/cfid}{cfid}
    package from the same authors (\citeproc{ref-RJ-2023-053}{Tikka, 2023})
    implements \texttt{ID*} (\citeproc{ref-shpitser2012idstar}{Shpitser \&
    Pearl, 2012}) and \texttt{IDC*}
    (\citeproc{ref-shpitser2012idstar}{Shpitser \& Pearl, 2012}). However,
    these packages are challenging to use and extend.

    Finally, \href{https://www.causalfusion.net}{CausalFusion} is a web
    application that implements many identification and estimation
    algorithms, but is neither open source, available for registration of
    new users, nor provides documentation.

    Causal inference remains an active research area where new algorithms
    are regularly published (see the recent review from Tikka et al.
    (\citeproc{ref-JSSv099i05}{2021})), but often without a reference
    implementation. We therefore implemented the \(Y_0\) Python package in
    order to address the need for open source implementations of existing
    algorithms as well as to provide a modular framework that can support
    the implementation of future algorithms and workflows.

    \section{Implementation}\label{implementation}

    \textbf{Probabilistic Expressions} \(Y_0\) implements an internal
    domain-specific language that can capture variables, counterfactual
    variables, population variables, and probabilistic expressions in which
    they appear. It covers the three levels of Pearl's Causal Hierarchy
    (\citeproc{ref-bareinboim2022}{Bareinboim et al., 2022}), including
    association \(P(Y=y \mid
    X=x^\ast)\), represented as \texttt{P(Y | \textasciitilde
    X)}, interventions \(P_{do(X=x^\ast)}(Y=y, Z=z)\), represented as
    \texttt{P[\textasciitilde X](Y, Z)} and counterfactuals
    \(P(Y_{do(X=x^\ast)}=y^\ast\mid X=x, Y=y)\), represented as
    \texttt{P(\textasciitilde Y @ \textasciitilde X | X, Y)}. Expressions
    can be converted to SymPy (\citeproc{ref-meurer2017sympy}{Meurer et al.,
        2017}) or LaTeX expressions and can be rendered in Jupyter notebooks.

    \textbf{Data Structure} \(Y_0\) builds on NetworkX
    (\citeproc{ref-hagberg2008networkx}{Hagberg et al., 2008}) to implement
    an (acyclic) directed mixed graph data structure, used in many
    identification algorithms, and the latent variable graph structure
    described by Evans (\citeproc{ref-evans2016simplification}{2016}). It
    includes a suite of generic graph operations, graph simplification
    workflows such as the one proposed by Evans, and conversion utilities
    for Ananke, CausalFusion, pgmpy, and causaleffect.

    \textbf{Falsification} \(Y_0\) implements several workflows for checking
    the consistency of graphical models against observational data. First,
    it implements D-separation (\citeproc{ref-Pearl_2009}{Pearl, 2009}),
    M-separation (\citeproc{ref-drton2004mseparation}{Drton \& Richardson,
        2004}), \(\sigma\)-separation
    (\citeproc{ref-forre2018sigmaseparation}{Forré \& Mooij, 2018}) that are
    applicable to increasingly more generic mixed graphs. Then, it
    implements a workflow for identifying conditional independencies
    (\citeproc{ref-Pearl1989}{Pearl et al., 1989}) and falsification
    (\citeproc{ref-eulig2023falsifyingcausalgraphsusing}{Eulig et al.,
        2023}). Finally, it provides a wrapper around \texttt{causaleffect}
    through \href{https://github.com/rpy2/rpy2}{\texttt{rpy2}} for
    calculating Verma constraints (\citeproc{ref-tian2012verma}{Tian \&
    Pearl, 2012}).

    \textbf{Identification} \(Y_0\) has the most complete suite of
    identification algorithms of any causal inference package. It implements
    \texttt{ID} (\citeproc{ref-shpitser2006id}{Shpitser \& Pearl, 2006}),
    \texttt{IDC} (\citeproc{ref-shpitser2007idc}{Shpitser \& Pearl, 2008}),
    \texttt{ID*} (\citeproc{ref-shpitser2012idstar}{Shpitser \& Pearl,
        2012}), \texttt{IDC*} (\citeproc{ref-shpitser2012idstar}{Shpitser \&
    Pearl, 2012}), surrogate outcomes (\texttt{TRSO})
    (\citeproc{ref-tikka2019trso}{Tikka \& Karvanen, 2019}),
    \texttt{tian-ID} (\citeproc{ref-tian2010identifying}{Tian \& Shpitser,
        2010}), transport (\citeproc{ref-correa2020transport}{S. Lee et al.,
        2020}), and counterfactual transport
    (\citeproc{ref-correa2022cftransport}{Correa et al., 2022}).

    \section{Case Study}\label{case-study}

\begin{figure}
        \centering
        \includegraphics[width=\textwidth,height=1.38889in]{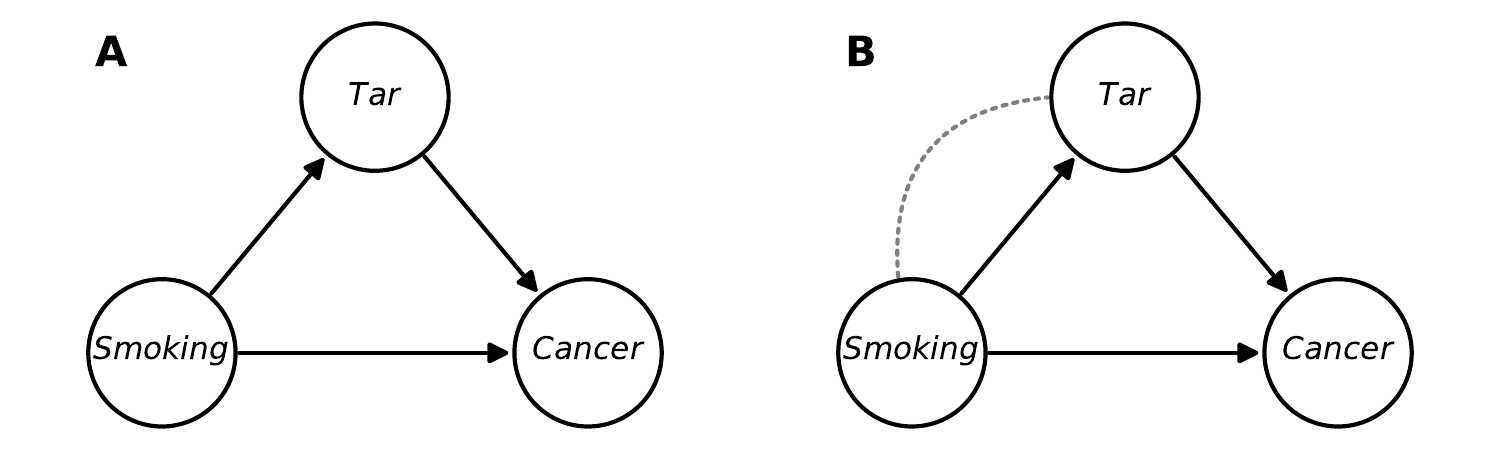}
        \caption{\textbf{A}) A simplified acyclic directed graph model
        representing prior knowledge on smoking and cancer and \textbf{B}) a
        more complex acyclic directed mixed graph that explicitly represents
        confounding variables.}\label{cancer}
    \end{figure}

    We present a case study regarding the effect of how smoking relates to
    cancer. First, we construct a graphical model (\autoref{cancer}A)
    representing the following prior knowledge:

    \begin{enumerate}
        \def\labelenumi{\arabic{enumi}.}
        \tightlist
        \item
        Smoking causes an accumulation of tar in the lungs.
        \item
        Accumulation of tar in the lungs increases the risk of cancer.
        \item
        Smoking also increases the risk of cancer directly.
    \end{enumerate}

    The identification algorithm (\texttt{ID})
    (\citeproc{ref-shpitser2006id}{Shpitser \& Pearl, 2006}) estimates the
    effect of smoking on the risk of cancer in \autoref{cancer}A as
    \(\sum_{Tar} P(Cancer | Smoking, Tar) P(Tar | Smoking)\). However, the
    model in \autoref{cancer}A is inaccurate because it does not represent
    confounders between smoking and tar accumulation, such as the choice to
    smoke tar-free cigarettes. Therefore, we add a \emph{bidirected} edge in
    \autoref{cancer}B. Unfortunately, \texttt{ID} can not produce an
    estimand for \autoref{cancer}B, which motivates the usage of an
    alternative algorithm that incorporates observational and/or
    interventional data. For example, if data from an observational study
    associating smoking with tar and cancer (\(\pi^{\ast}\)) and data from a
    randomized trial studying the causal effect of smoking on tar buildup in
    the lungs (\(\pi_1\)) are available, the surrogate outcomes algorithm
    (\texttt{TRSO}) (\citeproc{ref-tikka2019trso}{Tikka \& Karvanen, 2019})
    estimates the effect of smoking on the risk of cancer in
    \autoref{cancer}B as \(\sum_{Tar} P^{\pi^{\ast}}(Cancer |
    Smoking, Tar) P_{\text{Smoking}}^{{\pi_1}}(Tar)\). Code and a more
    detailed description of this case study can be found in the following
    \href{https://github.com/y0-causal-inference/y0/blob/main/notebooks/Surrogate\%20Outcomes.ipynb}{Jupyter
    notebook}.

    We provide a second case study demonstrating the transport
    (\citeproc{ref-correa2020transport}{S. Lee et al., 2020}) and
    counterfactual transport (\citeproc{ref-correa2022cftransport}{Correa et
    al., 2022}) algorithms for epidemiological studies in COVID-19 in this
    \href{https://github.com/y0-causal-inference/y0/blob/main/notebooks/Counterfactual\%20Transportability.ipynb}{Jupyter
    notebook}.

    \(Y_0\) has already been used in several scientific studies which also
    motivated its further development:

    \begin{itemize}
        \tightlist
        \item
        Mohammad-Taheri et al.
        (\citeproc{ref-mohammadtaheri2022experimentaldesigncausalquery}{2022})
        used \(Y_0\) to develop an automated experimental design workflow.
        \item
        Mohammad-Taheri et al. (\citeproc{ref-taheri2023adjustment}{2023})
        used \(Y_0\) for falsification against experimental and simulated data
        for several biological signaling pathways.
        \item
        Mohammad-Taheri et al. (\citeproc{ref-taheri2024eliater}{2024}) used
        \(Y_0\) and Ananke to implement an automated causal workflow for
        simple causal queries compatible with \texttt{ID}.
        \item
        Ness (\citeproc{ref-ness_causal_2024}{2024}) used \(Y_0\) as a
        teaching tool for identification and the causal hierarchy.
    \end{itemize}

    \section{Future Directions}\label{future-directions}

    There remain several high value identification algorithms to include in
    \(Y_0\) in the future. First, the cyclic identification algorithm
    (\texttt{ioID})
    (\citeproc{ref-forruxe92019causalcalculuspresencecycles}{Forré \& Mooij,
        2019}) is important to work with more realistic graphs that contain
    cycles, such as how biomolecular signaling pathways often contain
    feedback loops. Second, Missing data identification algorithms can
    account for data that is missing not at random (MNAR) by modeling the
    underlying missingness mechanism (\citeproc{ref-mohan2021}{Mohan \&
    Pearl, 2021}). Third, algorithms that provide sufficient conditions for
    identification in hierarchical causal models
    (\citeproc{ref-weinstein2024hierarchicalcausalmodels}{Weinstein \& Blei,
        2024}) would be useful for supporting causal identification in
    probabilistic programming languages, such as ChiRho
    (\citeproc{ref-chirho}{Bingham \& Witty, 2025}).

    Several algorithms noted in the review by Tikka et al.
    (\citeproc{ref-JSSv099i05}{2021}), such as generalized identification
    (\texttt{gID}) (\citeproc{ref-lee2019general}{S. Lee et al., 2019}) and
    generalized counterfactual identification (\texttt{gID*})
    (\citeproc{ref-correa2021counterfactual}{Correa et al., 2021}), can be
    formulated as special cases of counterfactual transportability.
    Therefore, we plan to improve the user experience by exposing more
    powerful algorithms like counterfactual transport through a simplified
    APIs corresponding to special cases like \texttt{gID} and \texttt{gID*}.
    Similarly, we plan to implement probabilistic expression simplification
    (\citeproc{ref-tikka2017b}{Tikka \& Karvanen, 2017b}) to improve the
    consistency of the estimands output from identification algorithms.

    It remains an open research question how to estimate the causal effect
    for an arbitrary estimand produced by an algorithm more sophisticated
    than \texttt{ID}. Agrawal et al.
    (\citeproc{ref-agrawal2024automated}{2024}) recently demonstrated
    automatically generating an efficient and robust estimator for causal
    queries more sophisticated than \texttt{ID} using ChiRho
    (\citeproc{ref-chirho}{Bingham \& Witty, 2025}), a causal extension of
    the Pyro probabilistic programming language
    (\citeproc{ref-bingham2018pyro}{Bingham et al., 2018}). Probabilistic
    circuits
    (\citeproc{ref-darwiche2022causalinferenceusingtractable}{Darwiche,
        2022}; \citeproc{ref-wang2023tractable}{Wang, 2023}) also present a new
    paradigm for tractable causal estimation. Such a generalization would
    enable the automation of downstream applications in experimental design.

    \section{Availability and Usage}\label{availability-and-usage}

    \(Y_0\) is available as a package on
    \href{https://pypi.org/project/y0}{PyPI} with the source code available
    at \url{https://github.com/y0-causal-inference/y0} under a BSD 3-clause
    license, archived to Zenodo at
    \href{https://zenodo.org/doi/10.5281/zenodo.4432901}{doi:10.5281/zenodo.4432901},
    and documentation available at \url{https://y0.readthedocs.io}. The
    repository also contains an interactive Jupyter notebook tutorial and
    notebooks for the case studies described above.

    \section{Acknowledgements}\label{acknowledgements}

    The authors would like to thank the German NFDI4Chem Consortium for
    support. Additionally, the development of \(Y_0\) has been partially
    supported by the following grants:

    \begin{itemize}
        \tightlist
        \item
        DARPA award HR00111990009
        (\href{https://www.darpa.mil/program/automating-scientific-knowledge-extraction}{Automating
        Scientific Knowledge Extraction})
        \item
        PNNL Data Model Convergence Initiative award 90001
        (\href{https://web.archive.org/web/20240518030340/https://www.pnnl.gov/projects/dmc/converged-applications}{Causal
        Inference and Machine Learning Methods for Analysis of Security
        Constrained Unit Commitment})
        \item
        DARPA award HR00112220036
        (\href{https://www.darpa.mil/program/automating-scientific-knowledge-extraction-and-modeling}{Automating
        Scientific Knowledge Extraction and Modeling})
        \item
        Award number DE-SC0023091 under the DOE Biosystems Design program
    \end{itemize}

    The authorship of this manuscript lists the primary contributors as the
    first and last authors and all remaining authors in alphabetical order
    by family name.

    \section*{References}\label{references}
    \addcontentsline{toc}{section}{References}

    \phantomsection\label{refs}

\end{document}